\documentclass[10pt,twocolumn,letterpaper]{article}

\usepackage{times}
\usepackage{epsfig}
\usepackage{graphicx}
\usepackage{amsmath}
\usepackage{amssymb}
\usepackage{multirow}
\usepackage{caption}
\usepackage{footmisc}

\usepackage[pagebackref=true,breaklinks=true,letterpaper=true,colorlinks,bookmarks=false]{hyperref}

\begin{document}

\title{Anticipation and next action forecasting in video: an end-to-end model with memory}

\date{}
\author{Fiora Pirri, Lorenzo Mauro, Edoardo Alati, Valsamis Ntouskos, Mahdieh Izadpanahkakhk, Elham Omrani\\
ALCOR Lab, Sapienza University of Rome, Italy\\
{\tt\small \{pirri,mauro,alati,ntouskos,izadpanahkakhk,omrani\}@diag.uniroma1.it}
}

\twocolumn[{%
\renewcommand\twocolumn[1][]{#1}%
\maketitle
\begin{center}
    \centering
    \includegraphics[width=.98\textwidth,height=5cm]{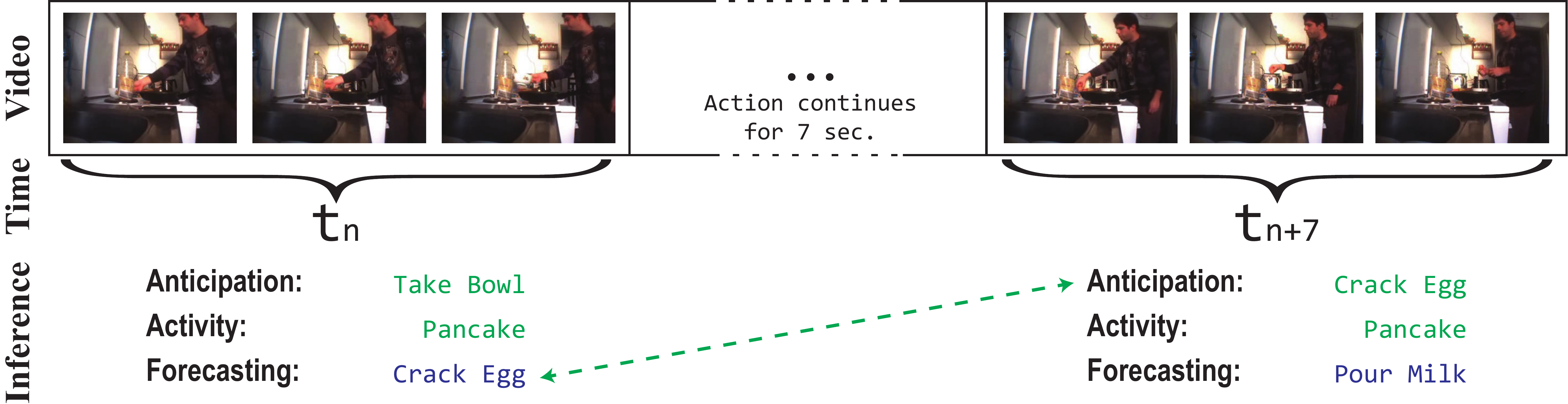}
    \captionof{figure}{Anticipation and next action forecasting with no time limits.}
   \label{fig:fig1}
\end{center}%
}]
\begin{center}
\centering
{\bf Abstract}
\end{center}
{\em Action anticipation and forecasting in videos do not require a hat-trick, as far as there are signs in the context to foresee how actions are going to be deployed. Capturing these signs is hard because the context includes the past.
We propose an end-to-end network for action anticipation and forecasting with memory, to both anticipate the current action and foresee the next one. Experiments on  action sequence datasets show excellent results indicating that training on histories with a dynamic memory can significantly improve forecasting performance.}

\graphicspath{{figs/}}

\section{Introduction}\label{sec:intro}
\begin{figure*}[t!]
 \centering
    \includegraphics[width=0.94\textwidth]{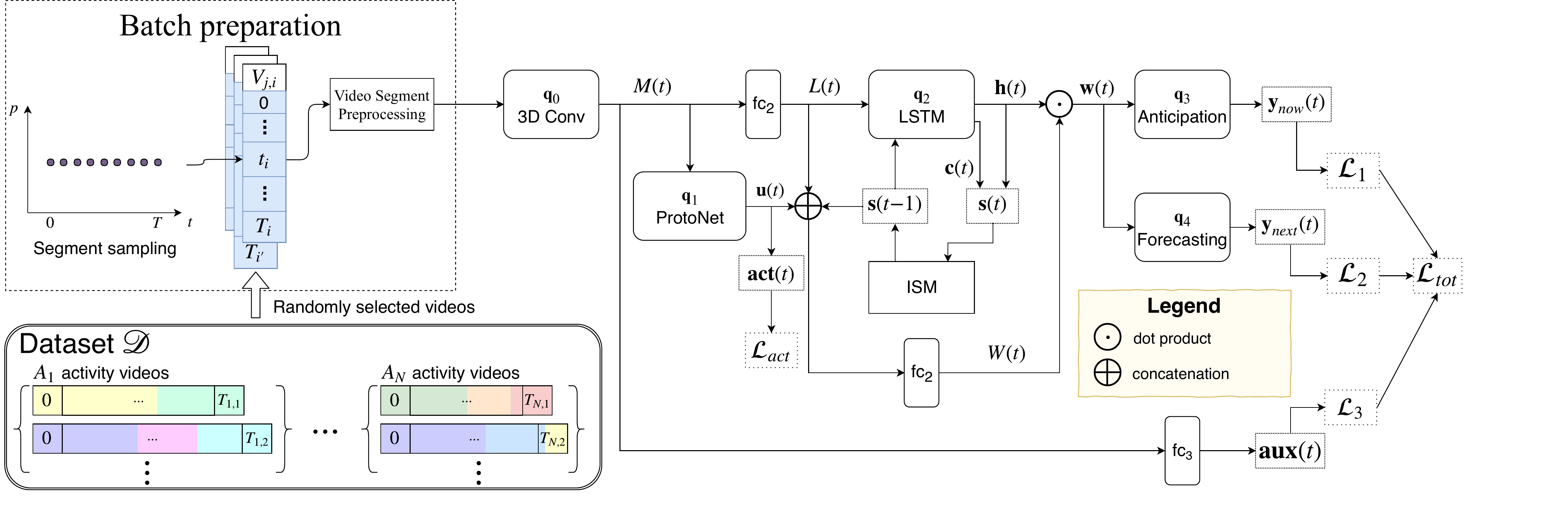}
    \caption{Schema of the Anticipation and Forecasting Network.}
   \label{fig:schema}
 \end{figure*}

There is nowadays an extraordinary amount of contributions to the challenging task of future events prediction, from a single image, a short video clip or a video.  A number of these contributions focus on anticipating actions recognition as earliest as possible \cite{hawkins2014,koppula2016,ma2016learning,shi2018}  obtaining significant results in reducing anticipation time. Others predict the features of a frame about seconds in the future \cite{lan2014,hawkins2014,vondrick2016,gao2017}. Other contributions focus on predicting motion trajectories, considering trajectory-based human activity \cite{kitani2012}, dense trajectories \cite{walker2016} or subsequent human movement paths \cite{vondrick2016}, which is quite relevant in team activities, especially sports.    Recently, \cite{yazanwill2018} have extended  long-term anticipation up to 5 minutes, assigning action labels to each frame in the temporal horizon of anticipation. 

A common denominator of the mentioned contributions is  the analysis of actions  beyond the visible frames, taking into account different scene features. Some have highlighted the relevance of objects for affordance \cite{koppula2016}, others the scene context, others the human motion or the analysis of past events. Proposed methods range from encoder-decoder models \cite{gao2017} to generative models \cite{liang2017dual}, to variational autoencoders \cite{walker2016}, and inverse reinforcement learning \cite{zeng2017}. 

Here we introduce a new variant of these ideas  conjugating  anticipation of the current action with forecasting the next unseen one.  Given few frames of a human activity  video, we can both anticipate the current action label and foresee the next one, independently of how long the current action will last. The idea is shown in Figure \ref{fig:fig1}. 

This variant of anticipation, requiring to foresee a completely unseen action, is useful in several contexts. It is  quite relevant to infer goals and intentions. In team sports, like soccer, it is useful for pursuing a specific game schema. In robotics it is useful for helping someone in accomplishing a task: knowing that after the current action a specific action might occur it would be possible for a robot to collect tools or predispose some machinery to provide help.

\section{ Overview and Contribution}
The main novelty of this work is the combination of anticipation of current action with forecasting the next unseen action by learning from the past,  using an accumulating memory.
Forecasting a next unseen action is challenging and it requires to anticipate to an early stage the recognition of the current action.  Further, it requires to discriminate plausible  future actions  filtering out those that are implausible. Finally, it requires to understand the context of the current action, and a relevant part of the context is the  past, for mastering action evolution.

To solve these problems we introduce a new end-to-end  Anticipation and Forecasting Network (AFN) with  memory to dynamically learn from past observations. The network schema is illustrated in Figure \ref{fig:schema}. AFN  anticipates the recognition in an untrimmed video from 6 frames scattered in one second,  examining several scenes and body elements, including motion and body pose for the current action and the action history.  We fix the  anticipation to one second to cope with the starting time uncertainty.  Further, a relevant component of AFN, besides the memory, is a ProtoNet \cite{snell2017protonets} predicting which activity an action belongs to.  Finally, based on the current action anticipation, AFN forecasts the next action, independently of the time horizon.

We illustrate our results on different baselines, considering all the newly introduced aspects. In particular, the anticipation is confronted with other methods on two  datasets used by several approaches, namely JHMDB \cite{Jhuang:ICCV:2013} and UCF101 \cite{UCF101}.  Forecasting is shown on four datasets Breakfast\cite{Kuehne12}, Charades \cite{sigurdsson2016hollywood}, CAD120 \cite{koppula2013dataset} and MPII cooking dataset \cite{rohrbach2012database}. We show that for both anticipation and forecasting, AFN outperforms the state of the art on these datasets.  In the lack of a specific metric, we have introduce appropriate measures, detailed in Section \ref{sec:results}.

\typeout{*********************** RELATED**********************}
\section{Related Work}\label{sec:related}
Since the early work of \cite{kitani2012} and \cite{ryoo2011human} a wealth of works focused on anticipation, prediction, and forecasting human actions, because is such a challenging and attractive research problem. So far the terms 
anticipation, prediction and forecasting have been used as fungible terms  to  estimate what shall occur in frames, which are not yet observed.  There is a difference, though. A first distinction is between the pixel level approaches   on the human motion  \cite{lan2014}, path  trajectories  \cite{jain2016structural,butepage2017deep,xu2018encoding} and image generation \cite{vondrick2017generating} and the approaches focusing on anticipating (predicting, forecasting) action labels. A general overview is given in   \cite{vondrick2016} and \cite{kong2018human}. 
A further difference is between anticipation and forecasting.
Here we exemplify the distinction as follows. While anticipation requires  a percentage of the current action to be observed, forecasting presumes that no frames of the foreseen action are observed. For this reason a number of datasets have  recently been proposed with long-term activity videos lasting minutes, where actors perform sequences of complex actions \cite{Kuehne12,damen2018dataset,sigurdsson2018,sigurdsson2016hollywood}.
 
\vskip 0.3 \baselineskip
\noindent
{\bf Action anticipation}
Action anticipation has been widely considered in the literature. Here we focus only on recent results on anticipation of action labels,  more details can be found in \cite{vondrick2016} and \cite{kong2018human}.
Recently,  ~\cite{rodriguez2018action} introduced tailored loss functions by applying dynamic images to a generative model for actions  anticipation. They benchmark their results mainly on JHMDB \cite{Jhuang:ICCV:2013}, UT-Interaction \cite{UT-Interaction-Data} and  UCF101  \cite{UCF101}. 
 Singh {\em et Al.} ~\cite{singh2017online} jointly evaluate  online spatio-temporal action localization and prediction via a single shot multi-box detector. Similarly, as ~\cite{rodriguez2018action}. They benchmark on UCF101 (though considering 24 actions) and JHMDB datasets. 
The  deep action tube network extended in \cite{singh2018predicting} has been designed in order to predict past, present, and future focusing on 10$\%$ of the video. It has been evaluated  on JHMDB dataset.
 Kong {\em et Al.} ~\cite{kong2018action} propose a  combined  CNN and  LSTM  along with a memory module in order to record ``hard-to-predict'' samples, they benchmark their results  on  UCF101 and on Sports-1M  \cite{karpathy2014large} datasets.
 In \cite{shi2018} an RNN with an RBF kernel is proposed to improve the performance of  feature mapping for the anticipation task. Their benchmarks are on JHMDB, UCF101 with 24 actions, and UT interaction. 
Lai {\em et Al.} ~\cite{lai2018global} optimize a global-local based temporal distance model and a two-stream network model. They benchmark their work on UCF11, the UCF YouTube Action Dataset \cite{liu2009recognizing} and on the HMDB with 51 actions. 
The work of \cite{aliakbarian2017encouraging} develops a multi-stage LSTM network  integrating both the context and action aware representations. They evaluated their work on the UCF101, UT-Interaction, and JHMDB datasets.
Considering the history trend, \cite{gao2017} applies a reinforced encoder-decoder to anticipate the next representation of an action. They evaluate their work on UCF101 and on TVSeries datasets.

Except \cite{singh2018predicting} and \cite{gao2017}, none of the above methods seems to focus on the past to anticipate actions. In fact, one of the contribution of our model is the addition of a memory component, which can deal with long sequences so as to learn from the action history both anticipation and forecasting. Also we confront with a larger set of actions from UCF101, JHMDB and other also for forecasting, reporting results significantly better than the cited ones, because the model proposed can learn from scattered sequences. 

\vskip 0.3 \baselineskip
\noindent
{\bf Forecasting future actions} Forecasting  unseen frames has been early introduced in \cite{lan2014} and \cite{vondrick2017generating}, the latter one promoting a significant amount of research on generating future frames based on generative adversarial networks. 
On the other hand, forecasting labels for unseen actions frames has not yet been widely developed, due to the fact that it is an ill-posed problem. As far as we know only   \cite{chakraborty2014context,mahmud2017joint} and  \cite{yazanwill2018} have faced forecasting for action labels. While \cite{mahmud2017joint}  use ground truth (in particular on MPII cooking activities dataset \cite{rohrbach2012database} and VIRAT dataset  \cite{oh2011large}) to forecast next action, we do not. Moreover, we can do both anticipation and forecasting on untrimmed video.
Similarly to our work  Farha {\em et Al.} \cite{yazanwill2018} have recently tackled the anticipation challenges by inferring current  and future actions via an RNN-HMM for anticipation and CNN for predicting the time horizon and the activity label. Like them, we use Breakfast \cite{Kuehne12} for forecasting and Charades \cite{sigurdsson2016hollywood} as a second dataset, in place of 50Salads \cite{stein2013combining}. The advantage of our approach is that we can combine anticipation and forecasting. Furthermore, as shown in the results section, we can foresee next action independently of time limits, and with greater accuracy.

\section{Anticipation and Forecasting Network}\label{sec:recognition}
The Action Forecasting Network (AFN) is the proposed end-to-end network. AFN aims  at both anticipating the current action  and forecasting the next action.
 To obtain the desired results we resort to several components.   To analyze the immediate previous frames we add  deep convolutional layers of a C3D network.  To establish the activity the action belongs to (in case activities are subdivided into actions) we add a ProtoNet. To take into account the history of the previous actions we add an RNN network with {\em lstm} cells and an internal dynamic memory. Finally, we use  several fully connected layers to connect the components and to both classify the current action and forecast the next one. As explained in \S \ref{sec:memory}, a relevant novelty of AFN is the ability to elaborate arbitrary long action sequences thanks to the introduction of an internal state memory  collecting the internal states of the RNN network components.
A visual representation of the network is illustrated in Figure \ref{fig:schema}.

\typeout{*************** Forecasting***************}

\subsection{Input}\label{sec:input}
For the video  datasets of activities taken into account, activities often allocate sequences of different actions, see Section \ref{sec:results} for details. Therefore,
given all the videos and activities taken from a chosen dataset,  the  input batch for a single run of the AFN network collects videos  both considering a number of activities and of actions. More specifically, we consider a dataset as 
\begin{equation}
{\mathcal D} =\{A_{j}{=}(a_{j,1},\ldots,a_{j,m}) | j{=}1,\ldots,N \mbox{ and } m{>}0\}
\end{equation} 
Where $A_{j}$ is an activity from a set of $N$ activities in the dataset ${\mathcal D}$ and the $a_{j,k}$ are the actions the activity $A_{j}$ might be decomposed into. For each activity $A_{j}$ there is a varying number of videos available. The input batch to AFN is formed  as follow. First we sample randomly from all the videos of all the activities in ${\mathcal D}$ a fixed number $\xi$ of videos, resulting in a sample of $K$,  $K<\xi$ of different activities. Further, we sample a segment  of $6$ RGB frames scattered on a time lapse of one second for each video. 
From the sampled RGB frames  we compute  two heat maps of the human pose (see \cite{cao2017realtime}):  one for the human joints and the other for the human limbs. We compute also the optical flow for each frame. This  information is stacked as four more channels in addition to the RGB ones. Finally,  all six frames, for each video, are collected into a single tensor as follows:
\begin{equation}\label{eq:input}
\begin{array}{l}
X_{j,i}(t_i){=} \pi(V_{j,i}(t_i)), \  t_i{\sim} Unif(0,T_i), i{=}1,{\ldots},\xi\\
\hspace*{15mm} T_i {=} length(V_{j,i}) \mbox{ in seconds}\ , j=1,{\ldots},K \\
\end{array}
\end{equation}
\noindent Here $\pi$ is the above described sampling and preprocessing of the $6$ frames sample,  $V_{j,i}$  denotes the $i$-th video of $A_j$, $j=1,\ldots,K$, and $t_i$ is the time index, randomly sampled from a uniform distribution in the interval $[0,T_i]$, with $T_i$ the length of the video $V_{j,i}$, in seconds. Therefore, an input batch is formed by the tensors $X_{j,i}$. In the following when we focus on a single tensor, we drop the indexes both from $X$ and from $t$.

\subsection{Internal State Memory}\label{sec:memory}
To allow AFN to take into account the history of the whole previous sequence we introduce an Internal State Memory (ISM), which stores the internal states of the RNN  network with {\em lstm} cells \cite{hochreiter1997} (indicated, resp., as ISM and $q_2$ in Figure \ref{fig:schema}). In this way every computation takes as input a single sample  and the internal state corresponding to the previous second. 
For every video in the dataset, a state vector $\mathbf{v} \in S^{T\times512}\subseteq {\mathbb R}^{T\times 512}$ 
is defined as:
\begin{equation}\label{eq:states}
\mathbf{v}=
  \begin{pmatrix}
   {\bf  s}(0)\\ 
    \vdots\\
    {\bf s}(T)
  \end{pmatrix}
\end{equation}
Here ${\bf s}(t)=({\bf h}(t)^{\top},\mathbf{c}(t)^{\top})^{\top}$, $t \in [0,T]$, are the $q_2$ states. Given $t$ we define ${\bf s}(t{-}1)$ as:
\begin{equation}
{\bf s}(t{-}1) =
\left\{
	\begin{array}{ll}
		\mathbf{0} & \mbox{if } J(t)=0 \\
		{\bf s}(t{-}1)^{J(t)} & \mbox{otherwise}
	\end{array}
\right.
\end{equation}
Here $J(t)$ denotes the number of times  that $t$ was sampled up to the current training step.

 Given the preprocessed input $X(t)$ and the internal state ${\bf s}(t{-}1)$ the $q_2$ component produces, beside the outputs, a new internal state ${\bf s}(t)^{J(t)}$. This new state is then used to update the states vector $\mathbf{v}$ (see eq. \ref{eq:states}).\\
The collection of  all the state vectors forms the ISM.
Note that at inference time, samples are taken sequentially.

\subsection{Anticipation and next action prediction}
The first component of AFN  are the 5 convolutional layers of C3D \cite{tran2015learning}, denoted   $q_{0}$, in Figure \ref{fig:schema}. This $q_{0}$ component computes  a feature tensor taking  $X(t)$ (see eq. \ref{eq:input}) as input:
\begin{equation}\label{eq:M}
M(t) =  q_{0}(X(t)).
\end{equation}
The tensor computed by $q_{0}$ has size ($7\times 7\times 512$) and is further reduced in size by a fully connected network $fc_{k}$, where $k$ indicates the network layers, here $k=2$. 
\begin{equation}\label{eq:L}
L(t) = fc_k(M(t))
\end{equation}
The next component is   $q_{2}$, introduced in  \S \ref{sec:memory}, which computes the vector:
\begin{equation}\label{eq:nextstate}
{\bf s}(t) = (\mathbf{h}(t)^{\top},\mathbf{c}(t)^{\top})^{\top} = q_{2}(L(t),{\bf s}(t{-}1))
\end{equation}
\noindent
Here ${\bf s}(t)$ and ${\bf s}(t{-}1)$ are $q_2$ states and the way they are selected from the ISM is described in the previous  paragraph \ref{sec:memory}.

At this point we introduce an advisory component $q_1$ supplying $q_2$ output with further information. This information is collected by a fully connected network, which learns the weights used to dynamically influence ${\bf h}(t)$.
More precisely, these weights are defined as follows:

\begin{equation}\label{eq:weights}
 W(t) = vec^{-1}(fc_k([L(t), {\bf s}(t{-}1), \mathbf{u}(t)]))
\end{equation}
\noindent
Here $L(t), {\bf s}(t{-}1), \mathbf{u}(t)$ are concatenated, $fc_k$ is a fully connected layer with $k=1$, $L(t)$ is the tensor obtained in eq. \ref{eq:L} and ${\bf u}(t)$ is the latent vector obtained by the component $q_1$. The component $q_1$  is a prototypical network, detailed in \S \ref{sec:pred}, which predicts the activity the current input belongs to, hence it provides relevant information, restricting the  search space for the final classification. The notation $vec^{-1}$ used in eq. \ref{eq:weights} indicates the inverse of the vectorization operator, namely $vec^{-1} : \mathbb{R}^{nm} \to \mathbb{R}^{m \times n}$.

\begin{figure}
 \centering
    \includegraphics[width=0.45 \textwidth]{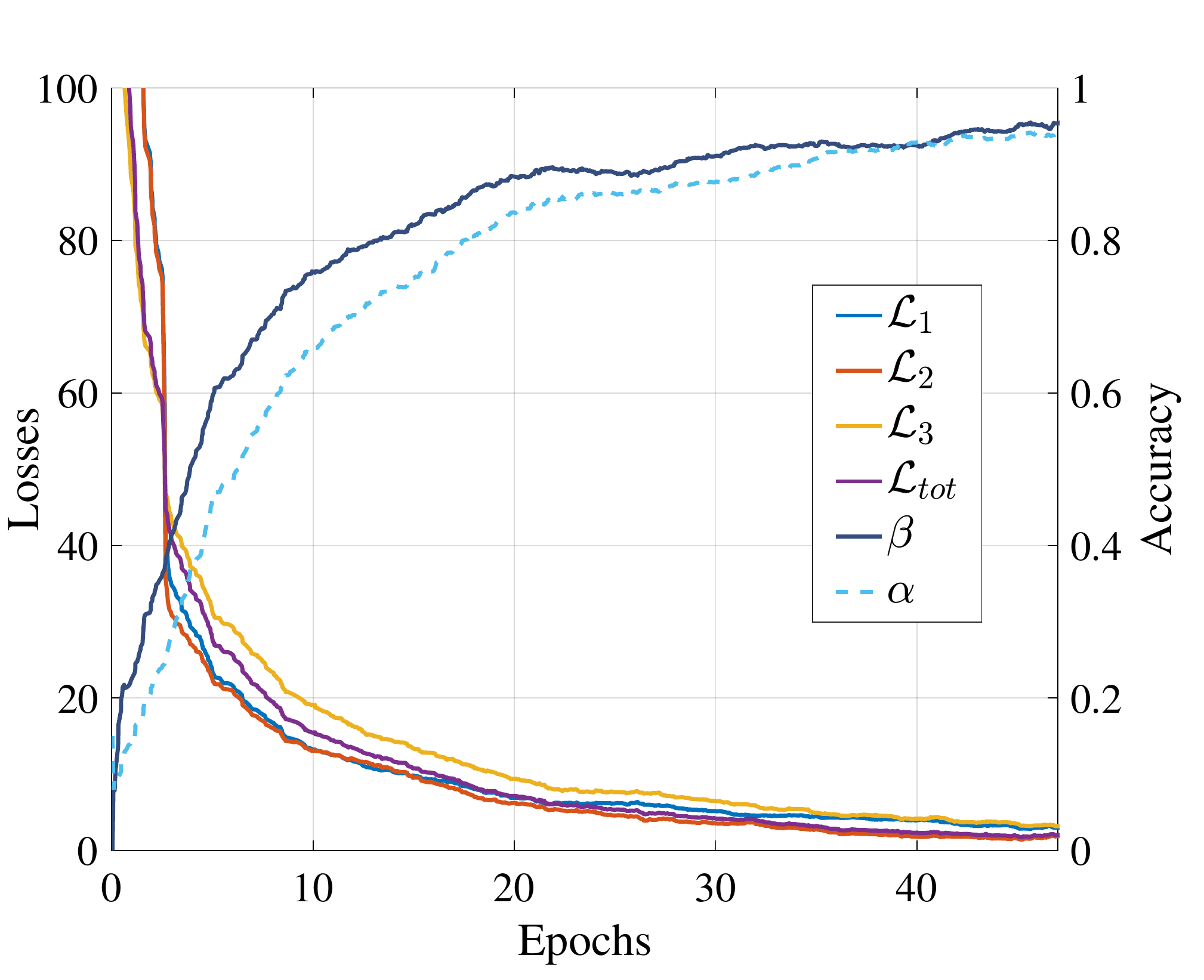}
    \caption[]{Training loss and accuracy per  epochs.}
   \label{fig:accuracies}
 \end{figure}

The weight matrix $W$ of size $256 \times 256$ is finally multiplied by the $q_2$ output ${\bf}(t)$, obtaining the vector ${\bf w}(t)$:
\begin{equation}
{\bf w}(t) = W(t){\bf h}(t)
\end{equation}
The last classification component is $q_{3}$, anticipating the current action as follows:
\begin{equation}\label{eq:finalActual}
{\bf y}_{now}(t) = q_{3}({\bf w}(t)) = \sigma(fc_k(\max(0, {\bf w}(t) + {\bf b}_{now})))
\end{equation}
where $\sigma(x_j)=  e^{x_j}/\sum\limits_{j=0}^K e^{x_j}$ is the softmax, ${\bf b}_{now}$ is the bias, and $K$ are the action classes.

Finally, ${\bf y}_{now}(t)$ and ${\bf w}(t)$ are elaborated by a secondary fully connect layer $fc_k$ with $k=1$ to predict the next action ${\bf y}_{next}$. This is possible since the  vector ${\bf w}(t)$ keeps both the information of the current input $X(t)$ (see eq. \ref{eq:input}) and the information of the  previous actions, due to the interaction between the memory and $q_2$. The current action classification vector adds up these information  to predict the next action:
\begin{equation}\label{eq:next}
{\bf y}_{next}(t) = \sigma(fc_k([\max(0, {\bf w}(t) + {\bf b}_{next}), {\bf y}_{now}(t)]))
\end{equation}

To speed up the contribution of the convolutional layers of $q_0$ to the optimization, we define a loss function taking into account a direct $M(t)$  classification (see eq. \ref{eq:M}):
\begin{equation}
{\bf aux}(t) = \sigma(fc_k(M(t)))
\end{equation}
\noindent
Here $fc_k$ is a fully connected layer with $k=3$.

Having defined the output of  AFN  we illustrate the cross entropy loss function on the three outputs, namely $\mathcal{L}_1$ on ${\bf y}_{now}$, $\mathcal{L}_2$ on ${\bf y}_{next}$ and $\mathcal{L}_3$ on ${\bf aux}$.

\begin{equation}\label{eq:L_actual}
\mathcal{L}_1(t) = - \sum_{i} {\bar{y}_{i}(t)} \log ({{y_{now}}_i(t)})\\
\end{equation}
\begin{equation}\label{eq:L_next}\
\mathcal{L}_2(t) = - \sum_{i} {{\hat{y}}_{i}(t)} \log ({{y_{next}}_i(t)})
\end{equation}
\begin{equation}\label{eq:L_aux}
\mathcal{L}_3(t) = - \sum_{i} {\bar{y}_{i}(t)} \log ({{aux}_i(t)})
\end{equation}

Here $\bar{y}$ and $\hat{y}$ are the target vectors, respectively for the current action and the next, and $i$ is the $i$-th element of the corresponding vector.
The total loss is obtained by combining the above loss function as follow:

\begin{equation}\label{eq:L_tot}
\begin{split}
    \mathcal{L}_{tot}(t) & = \alpha(t^*)((1-\beta(t^*))\mathcal{L}_1(t)+\beta(t^*)\mathcal{L}_2(t)) \\ 
    & + (1-\alpha(t^*))\mathcal{L}_3(t) 
\end{split}
\end{equation}

Here $t^*$ is the previous training step, $\alpha$ is the accuracy measured on the auxiliary classification ${\bf aux}$, and $\beta$ is the accuracy computed on the current classification ${\bf y}_{now}$. 
The accuracy parameters $\alpha$ and $\beta$ combine $\mathcal{L}_1$, $\mathcal{L}_2$ and $\mathcal{L}_3$ dynamically, according to the performances reached by the network at the previous training step $t^*$.
\noindent
The following equation illustrates how $\mathcal{L}_{tot}$ changes in four  key steps of the training process,  where $\epsilon$ is a small positive value:
\begin{equation}\label{eq:loss_dynamic}
\mathcal{L}_{tot}(t) =
\left\{
	\begin{array}{ll}
		\mathcal{L}_3(t) 								  	& \mbox{if } \alpha \in [0, \epsilon]\mbox{ and }\beta \in [0, \epsilon] \\
		\\
		\frac{1}{2}(\mathcal{L}_1(t) + \mathcal{L}_3(t)) 	& \mbox{if } \alpha \in [\frac{1}{2} - \epsilon, \frac{1}{2} + \epsilon]\mbox{ and }\\
															& \beta \in [0, \epsilon] \\
															\\
		\frac{1}{2}(\mathcal{L}_2(t) + \mathcal{L}_1(t))	& \mbox{if } \alpha \in [1 - \epsilon, 1]\mbox{ and }\\
															& \beta \in [\frac{1}{2} - \epsilon, \frac{1}{2} + \epsilon] \\
															\\
		\mathcal{L}_2(t)									& \mbox{if } \alpha \in [1 - \epsilon, 1]\mbox{ and }\\
															& \beta \in [1 - \epsilon, 1] \\
	\end{array}
\right.
\end{equation} 

The parameters $\alpha$ and $\beta$ in eq. \ref{eq:L_tot} enforce the optimization of the convolution component $q_{0}$ in the early training  steps, when both $\alpha$ and $\beta$ are close to $\epsilon$ (first line in eq. \ref{eq:loss_dynamic}). On the other hand as they both reach max values (last line in eq. \ref{eq:loss_dynamic}), they push the optimization toward ${\bf y}_{next}$. In the intermediate steps (second and third line in eq. \ref{eq:loss_dynamic}) $\alpha$ and $\beta$ favor the optimization of ${\bf y}_{now}$. Clearly, the transitions among these steps are smooth and we can note that $\mathcal{L}_3$ optimization is gradually ignored as $\alpha$ increases. Similarly, $\mathcal{L}_1$ optimization is steadily disregarded as soon as $\beta$ increases. The graph in Figure \ref{fig:accuracies} illustrates the above considerations.

\subsection{Activities embedding}\label{sec:pred}
\begin{figure}[t!]
 \centering
    \includegraphics[width=0.98 \columnwidth]{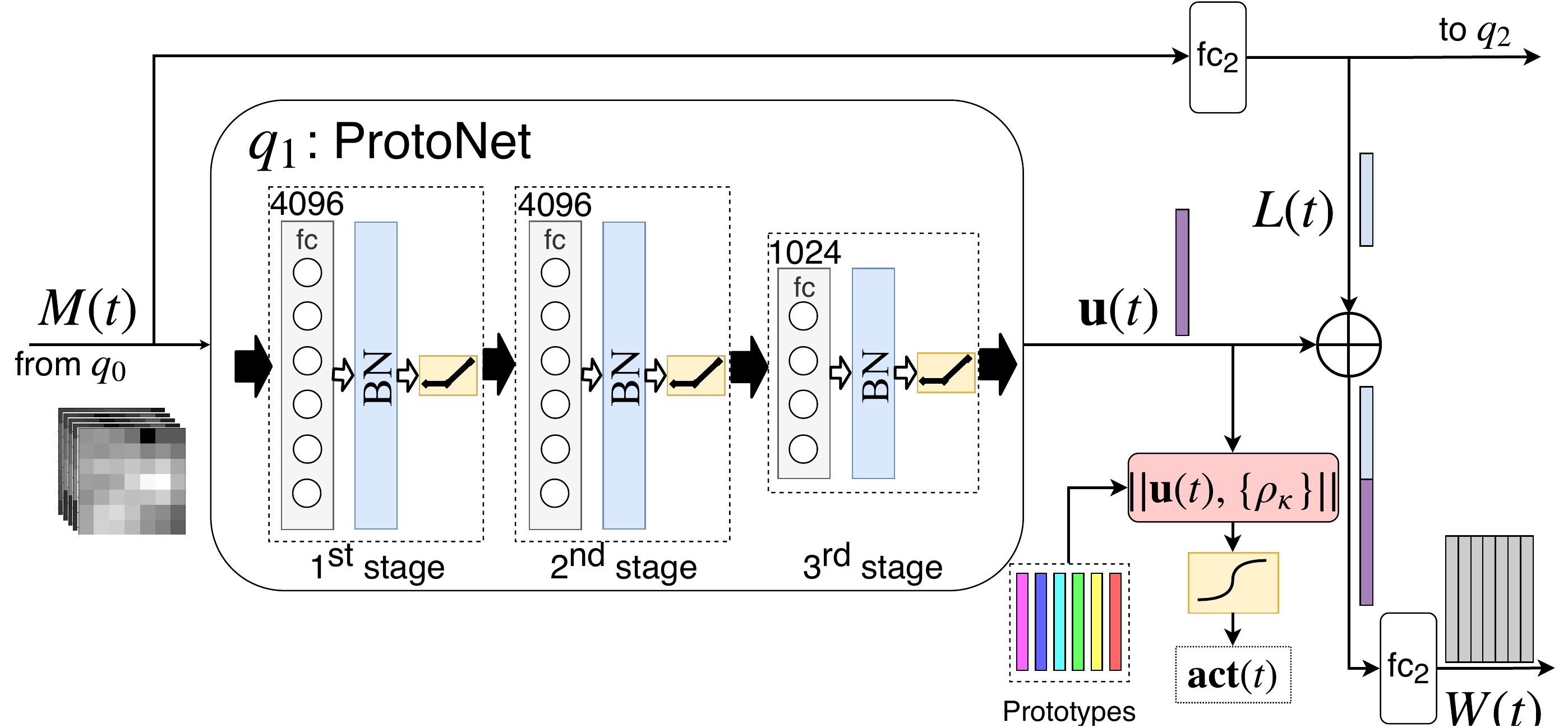}
    \caption[]{Prototypical Network for activity classification and advising.}
   \label{fig:protonet}
 \end{figure}

 The $q_1$ component of AFN is an activity encoder which produces an embedding representative of the activity performed. The produced embedding provides an essential context for forecasting the action that follows by predicting the activity class the observed tensor belongs to. 
 
 In order to make the network learn a representative embedding, we consider a Prototypical Network (ProtoNet) \cite{snell2017protonets} comprising three stages each formed by a fully connected layer, a batch normalization layer and a ReLU activation function (see Figure~\ref{fig:protonet}). The fully connected layers are of size $4096$, $4096$ and $1024$, respectively. The network receives sample tensors $M(t)$ from $q_0$ and predicts the activity which contextualizes the performed actions via the embedding described below. 
 
The input batch contains $\xi$ videos corresponding to $K$ activities. Reintroducing the indices used in  \S \ref{sec:input}, we define $\mathcal{M}_j=\{M_{j,1},\ldots,M_{j,\zeta}\}$ as a collection of feature tensors corresponding to the $\zeta$ samples of activity $A_j$ contained in the batch. Similarly to \cite{snell2017protonets}, we assume that there exists a latent space where the embedded vectors $\mathbf{u}=f_\theta(M)$, with $M\in\mathcal{M}_j$, cluster around the corresponding activity prototype $\rho_j$. In our context, letting $\theta$ be the parameters of component $q_1$, the network training approximates this optimal embedding function $f_\theta(\cdot)$. In this way, $q_1$ is used to compute embedded vectors which contain context about the performed activity and therefore provide advised information for forecasting the next action.

 Specifically, two disjoint subsets $C_j$ and $Q_j$ are sampled randomly from the collection $\mathcal{M}_j$. 
 Letting $|C_j|$ denote the cardinality of $C_j$, the prototype of activity $A_j$ is defined as the sample mean of the embedded vectors corresponding to the samples contained in $C_j$, namely 
\begin{equation}
\rho_{j} = \frac{1}{|C_j|}\sum_{M\in C_j} f_\theta(M).
\end{equation} 
 The likelihood of an instance $M_q \in Q_j$ belonging to activity $j$ is then given by
 \begin{equation}\label{eq:protolhood}
 p(j|M_q,\theta) = \frac{\gamma_j\exp(-\|f_\theta(M_q) {-} \rho_j \|)}{\sum_{\kappa}\gamma_\kappa\exp(-\|f_\theta(M_q){-}\rho_{\kappa}\|)}.
 \end{equation}
 Here, $\gamma_j$ are learnable parameters that represent the relative weight of each activity cluster in the latent space.
 
  Considering now that the class of $M_q$ is $j$, the parameters $\theta$ are computed via Stochastic Gradient Descent by minimizing the loss 
  \begin{align}
 \mathcal{L}_{act} =& 
 \sum_{j{=}1}^{K}\sum_{M {\in} Q_j}\bigg\{\|f_\theta(M) {-} \rho_j \|-\log\gamma_j  \nonumber\\
 &+ \log\Big( \sum_{\kappa}\gamma_\kappa \exp({-}\|f_\theta(M){-}\rho_{\kappa}\|)
\Big)\bigg\},
 \end{align}
  which corresponds to the negative log of (\ref{eq:protolhood}) over the  batch. 
 
 The latent vector $\mathbf{u}=f_\theta(M)$ computed by the ProtoNet is then concatenated to the output of $fc_2$, according to eq.~\ref{eq:weights}, to provide context for forecasting the next action.

\typeout{*************** Results***************}
\section{Results}\label{sec:results}
\begin{figure*}[t!]
 \centering
    \includegraphics[width=0.95 \linewidth]{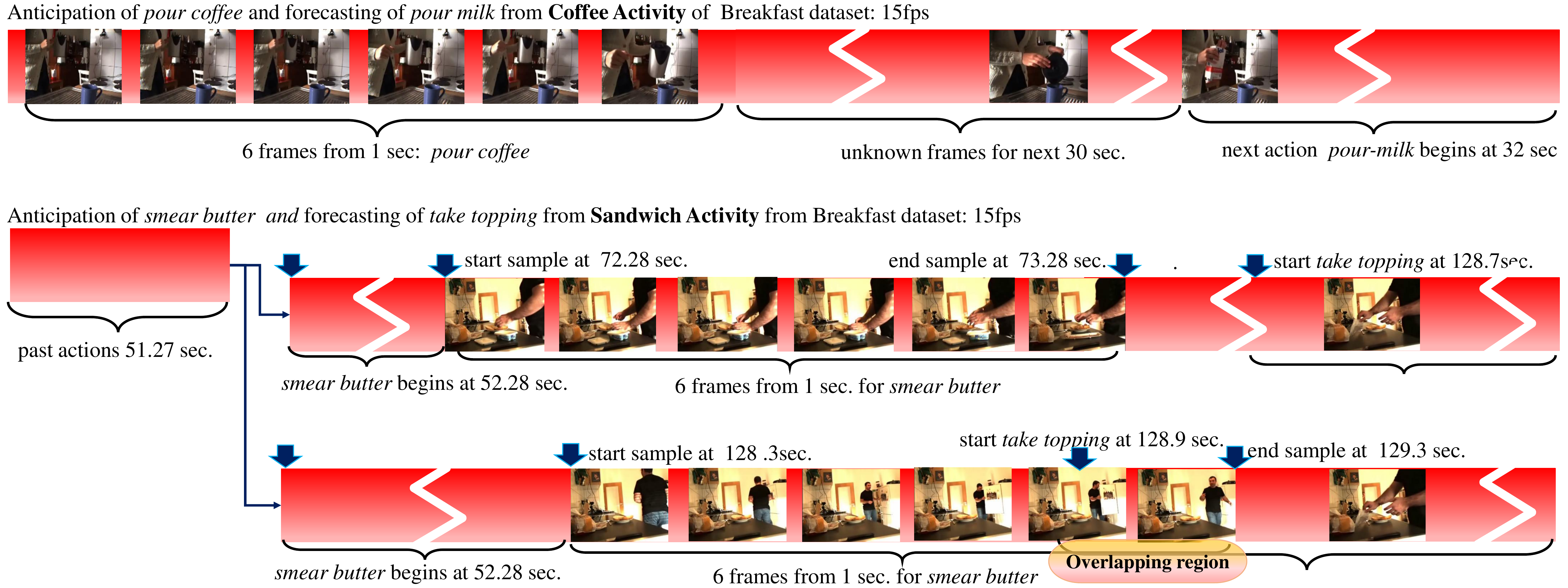}
    \caption{The  schema illustrates the effects of sampling 6 frames from 1 sec. at any point in a video, at inference time. The general idea is shown in the first strip: anticipation here is $3.3\%$ of the action {\em pour coffee}  and  the next action is predicted 30 sec. in advance. Below, we observe two cases that might occur when jumping into a video at a random time point, here 72.28 sec. after the video start.  In the second strip, the current action {\em smear butter} is anticipated, and the next action {\em take topping} starting 56 sec. after it, is forecast.  In the third stripe, the sample is taken on an interval overlapping  {\em smear butter} and {\em take topping}, in this case   neither anticipation nor forecasting occurs. Details are given in \S \ref{ssec:eval}.}\label{fig:tempiazioni}
 \end{figure*}

\noindent
{\bf Overview.}
 By action {\em anticipation}  we intend action labeling given  just a few observed frames. Slightly differently from the literature  (e.g. \cite{shi2018}) AFN  samples 6  frames, out of a second,   at any point in a video, not just at the beginning of the action. In fact, we do not rely on one shot actions nor on trimming at inference time.  In other words,  we consider  videos of any length  without taking into account actions start and end. 
By  action {\em forecasting}, we intend labeling an action when no frames   have been yet observed, and only the previous action has been anticipated, according to the above definition of {\em anticipation}.  Figure \ref{fig:tempiazioni} illustrates anticipation and forecasting.   

We consider two types of activities, according to the datasets taken into account (see next paragraph). The first type,  used for anticipation, collects activities not further subdivided into action sequences. The second type, suitable for forecasting future actions, is split into actions forming sequences, which can be represented as directed (not acyclic) graphs. Because AFN both anticipates and forecasts the next action, to confront with the state of the art we take into account both dataset types, in so showing the extreme flexibility of its architecture.

\subsection{Datasets Selection}\label{ssec:dataset}
For {\em action anticipation} we have selected the two datasets common to most of the recent  works on early action recognition, namely {\bf JHMDB} and  {\bf UCF101} datasets. JHMDB \cite{Jhuang:ICCV:2013} is composed by 928 videos of 21 selected actions out of the 51 of the original HMDB \cite{kuehne2011hmdb}.  
\textbf{UCF101} \cite{UCF101} has 101 action categories  for 430 hours of videos.

Experiments on {\em action forecasting}, are done on four datasets:  {\bf Breakfast} \cite{Kuehne12},  {\bf Charades} \cite{sigurdsson2016hollywood},  {\bf CAD120} \cite{koppula2013dataset} and  {\bf MPII cooking dataset} \cite{rohrbach2012database}.

The Breakfast \cite{Kuehne12}  dataset has 10 activity categories over 48 actions. The dataset consists of 712 videos taken by 52 different actors. 
\noindent
Charades  dataset \cite{sigurdsson2016hollywood} combines 40 objects and 30 actions in 15 scenes generating a number of 157 action classes  for 9848 videos. To experiment on Charades dataset, we defined the contexts as the activities, slightly modifying the temporal annotations, since we do not consider objects. For example, living-room-activity collects the actions performed in a living room.

MPII cooking activities dataset \cite{rohrbach2012database} has 65 action classes for 14 dishes preparation activities, performed by 12 participants. There are 44 videos for about 8 hours of videos.

\subsection{Implementation Details} The whole set of experiments has been performed on two computers, both equipped with an i9 Intel processor and 128GB of RAM, and with $4$ NVIDIA Titan V.  AFN is developed using Tensorflow $1.11$. 

\noindent
{\bf Protocols} For training and testing our system, we used in turn all the datasets, fixing $60\%$ for training, $30\%$ for testing and $10\%$ for validation, where not already defined by the dataset. We did not select  the actors to separate training, validation and test, since in some datasets not all actions are performed by each actor. For optimization we used Adam \cite{kingma2014adam} and gradient clipped to $1$.  We  trained the model with different batch sizes, mainly to cope with the ProtoNet component of the network, $q_1$. The final size was $25$, due to the large tensors (see eq. \ref{eq:input}). The image size at input is $112{\times} 112$. We  used a dynamic learning rate, starting with   $10^{-4}$,  decreasing it of a factor of $0.9$ every $3k$ steps. We used  dropout $0.6$ only at the convolutional layers of $q_0$.  The training speed is roughly 11 steps per sec. Having $45$ Epochs, each of  $10k$ steps, the overall training time is about $43k$ seconds.
\begin{table}[t!]
\caption {Comparison on Anticipation} \label{tab:anticipation} 
\begin{minipage}{\columnwidth}
\resizebox{\columnwidth}{!}{
\begin{tabular}{|l|l|l|l|}
\hline
\multicolumn{4}{|c|}{\textbf{UCF101}}                                                                                                  \\ \hline
\textbf{Video}      & \textbf{\%}          & \textbf{Framework}                                                 & \textbf{Accuracy} \\ 
\textbf{length}      &          &                                               &  \\ 
\hline
\multirow{7}{*}{2s}  & \multirow{7}{*}{0.5} & mem-LSTM \cite{kong2018action}                                        & 83.37\%           \\ \cline{3-4} 
                     &                      & DeepSCN \cite{kong2017deep}                                           & 85.46\%           \\ \cline{3-4} 
                     &                      & A+AF \cite{singh2017online} \footnote{\label{ucf24}UCF101 (24 actions)}            & $\sim 87\%$       \\ \cline{3-4} 
                     &                      & fm+RBF+GAN+Inception V3 \cite{shi2018} \footref{ucf24}                & 98,00\%           \\ \cline{3-4} 
                     &                      & global-local LSTM \cite{lai2018global} \footnote{\label{ucf11}UCF11} & 74.30\%           \\ \cline{3-4} 
                     &                      & Two-Stream Network \cite{lai2018global} \footref{ucf11}               & 90.50\%           \\ \cline{3-4} 
                     &                      & \textbf{AFN}                                                          & \textbf{98.30\%}  \\ \hline
\multirow{4}{*}{5s}  & \multirow{4}{*}{0.2} & global-local LSTM \cite{lai2018global} \footref{ucf11}                & 70.20\%           \\ \cline{3-4} 
                     &                      & Two-Stream Network \cite{lai2018global} \footref{ucf11}               & 89.3\%            \\ \cline{3-4} 
                     &                      & A+AF\cite{singh2017online} \footref{ucf24}                            & $\sim 83\%$       \\ \cline{3-4} 
                     &                      & \textbf{AFN}                                                          & \textbf{93.73\%}  \\ \hline
\multirow{8}{*}{10s} & \multirow{8}{*}{0.1} & global-local LSTM \cite{lai2018global} \footref{ucf11}                & 68.50\%           \\ \cline{3-4} 
                     &                      & Two-Stream Network \cite{lai2018global} \footref{ucf11}               & 87.5\%            \\ \cline{3-4} 
                     &                      & Static and Dynamic CNN \cite{rodriguez2018action} \footref{ucf24}     & 89.30\%           \\ \cline{3-4} 
                     &                      & DeepSCN \cite{kong2017deep}                                           & 44.31 \%          \\ \cline{3-4} 
                     &                      & mem-LSTM \cite{kong2018action}                                        & 51.02\%           \\ \cline{3-4} 
                     &                      & RED \cite{gao2017}                                                    & 37.5\%            \\ \cline{3-4} 
                     &                      & Deep K-3 \cite{vondrick2016}                                          & 43,6\%            \\ \cline{3-4} 
                     &                      & \textbf{AFN}                                                          & \textbf{91.23\%}  \\ \hline
\multicolumn{4}{|c|}{\textbf{JHMDB}}                                                                                                    \\ \hline
\multirow{4}{*}{2s}  & \multirow{4}{*}{0.5} & TP Net \cite{singh2018predicting}\footnote{It uses atomic actions}    & 74.1\%            \\ \cline{3-4} 
                     &                      & Multi-Stage LSTM \cite{aliakbarian2017encouraging}                    & 58.0\%            \\ \cline{3-4} 
                     &                      & A+AF \cite{singh2017online}                                           & $\sim 63\%$       \\ \cline{3-4} 
                     &                      & \textbf{AFN}                                                          & \textbf{85.4\%}   \\ \hline
\multirow{6}{*}{5s}  & \multirow{6}{*}{0.2} & Static and Dynamic CNN \cite{rodriguez2018action}                     & 61.0\%            \\ \cline{3-4} 
                     &                      & TP Net \cite{singh2018predicting}                                     & 74.8\%            \\ \cline{3-4} 
                     &                      & Multi-Stage LSTM \cite{aliakbarian2017encouraging}                    & 55.0\%            \\ \cline{3-4} 
                     &                      & A+AF\cite{singh2017online}                                            & $\sim 60\%$       \\ \cline{3-4} 
                     &                      & fm+RBF+GAN+Inception V3 \cite{shi2018}                                & 73\%              \\ \cline{3-4} 
                     &                      & \textbf{AFN}                                                          & \textbf{81.6\%}   \\ \hline
\end{tabular}
}
\end{minipage}
\end{table}
\begin{table}[h!]
\caption{Comparison on Action Forecasting}\label{tab:Bonn}
\centering
\begin{tabular}{|l|c|c|}
\hline
\textbf{Prediction}                                                                                         & \multicolumn{2}{c|}{\textbf{10\% (Next Action)}}                              \\ \hline
\textbf{Observation}                                                                                        & \textbf{20\%}                         & \textbf{30\%}                         \\ \hline
RNN-based Anticipation \cite{yazanwill2018}                                                                 & 18.11\%                               & 21.64\%                               \\ \hline
\begin{tabular}[c]{@{}l@{}}RNN-based Anticipation \\ (Ground Truth given) \cite{yazanwill2018}\end{tabular} & 60.35\%                               & 61.45\%                               \\ \hline
CNN-based Anticipation \cite{yazanwill2018}                                                                 & 17.90\%                               & 22.44\%                               \\ \hline
\begin{tabular}[c]{@{}l@{}}CNN-based Anticipation\\ (Ground Truth given) \cite{yazanwill2018}\end{tabular}  & 57.97\%                               & 60.32\%                               \\ \hline
AFN                                                                                                         & \multicolumn{1}{l|}{\textbf{91.09\%}} & \multicolumn{1}{l|}{\textbf{93.27\%}} \\ \hline
\end{tabular}
\end{table}
 \begin{table}[h!]
\caption{Comparison on next Action Forecasting I}\label{tab:MPII1}
\resizebox{\columnwidth}{!}{
\begin{tabular}{|l|l|l|l|}
\hline
\multicolumn{4}{|l|}{\textbf{MPII Cooking  Dataset}}                                                                                                                           \\ \hline
\textbf{Framework}                                             & \textbf{Precision} & \textbf{Recall} & \textbf{\begin{tabular}[c]{@{}l@{}}Accuracy \\ (Top-1)\end{tabular}} \\ \hline
Activity Label Prediction \cite{mahmud2017joint} & 70.7\%             & 66.5\%          & 80.1\%                                                               \\ \hline
AFN                                                               & \textbf{78.5\%}    & \textbf{74.6\%} & \textbf{86.2\%}                                                      \\ \hline
\end{tabular}

}

\end{table} 

\begin{table}[h!]
\caption{Comparison on next Action Forecasting II}\label{tab:MPII2}
\resizebox{\columnwidth}{!}{
\begin{tabular}{|l|l|l|l|l|l|}
\hline
\multicolumn{1}{|c|}{\textbf{Framework}} & \multicolumn{1}{c|}{\textbf{\begin{tabular}[c]{@{}c@{}}Sequence\\ length\\ 2\end{tabular}}} & \multicolumn{1}{c|}{\textbf{\begin{tabular}[c]{@{}c@{}}Sequence\\ length\\ 3\end{tabular}}} & \multicolumn{1}{c|}{\textbf{\begin{tabular}[c]{@{}c@{}}Sequence\\ length\\ 5\end{tabular}}} & \multicolumn{1}{c|}{\textbf{\begin{tabular}[c]{@{}c@{}}Sequence \\ length\\ 7\end{tabular}}} & \multicolumn{1}{c|}{\textbf{\begin{tabular}[c]{@{}c@{}}Sequence\\ length\\ 9\end{tabular}}} \\ \hline
ALP \cite{mahmud2017joint}                  & 78.8\%                                                                                      & 80.1\%                                                                                      & 79.2\%                                                                                      & 77.8\%                                                                                       & 77.2\%                                                                                      \\ \hline
\textbf{AFN}                                & \textbf{89.6\%}                                                                             & \textbf{92.7\%}                                                                             & \textbf{92.8\%}                                                                             & \textbf{93.3\%}                                                                              & \textbf{94.1\%}                                                                             \\ \hline
\end{tabular}
}

\end{table}
\begin{table}[h!]
\caption{Ablation with respect to next action forecasting}\label{tab:abx}
\resizebox{\columnwidth}{!}{
\begin{tabular}{|l|l|l|l|}
\hline
\textbf{\begin{tabular}[c]{@{}l@{}}Removed\\ Component\end{tabular} }        & \textbf{Dataset} & \textbf{\begin{tabular}[c]{@{}l@{}}Current Action\\Accuracy\end{tabular}} & \textbf{\begin{tabular}[c]{@{}l@{}}Next Action\\Accuracy\end{tabular}} \\ \hline
\multirow{3}{*}{$q _{1}$}         & Breakfast        & 85.06\%                                                               & 80.54\%                                                            \\ \cline{2-4} 
                                  & Charades         & 81.23\%                                                               & 78.65\%                                                            \\ \cline{2-4} 
                                  & CAD-120          & 82.34\%                                                               & 79.89\%                                                            \\ \hline
\multirow{3}{*}{$q_{4}$}          & Breakfast        & 87.74\%                                                               & \multicolumn{1}{c|}{---}                                           \\ \cline{2-4} 
                                  & Charades         & 84.34\%                                                               & \multicolumn{1}{c|}{---}                                           \\ \cline{2-4} 
                                  & CAD-120          & 84.75\%                                                               & \multicolumn{1}{c|}{---}                                           \\ \hline
\multirow{3}{*}{$\mathcal{L}_{3}$} & Breakfast        & 79.12\%                                                               & 76.23\%                                                            \\ \cline{2-4} 
                                  & Charades         & 75.59\%                                                               & 71.98\%                                                            \\ \cline{2-4} 
                                  & CAD-120          & 77.03\%                                                               & 74.29\%                                                            \\ \hline
\end{tabular}
}
\end{table}

\subsection{Comparison with the State-of-the-Art}
The comparison with the state of the art has been done for both anticipation and forecasting. Note that, as AFN uses softmax, accuracy is simply frequency count of  correct predictions out of all predictions.
In order to make the confrontation as fair as possible for all the works taken into account, we have considered different measures for both anticipation and forecasting.

\vskip 0.3 \baselineskip
\noindent
{\bf Action anticipation} 
For action anticipation we considered the datasets JHMDB \cite{Jhuang:ICCV:2013} and  UCF101\cite{UCF101} as discussed in \S \ref{ssec:dataset}. Results are shown in Table \ref{tab:anticipation}. Anticipation values are defined as:
\begin{equation}
\Delta^{-} = 1/\mu_{\ell}(V)  
\end{equation}  
where $\Delta^{-}\in \{0.1,0.2,0.5\}$ is the anticipation value and $\mu_{\ell}(V)$ is the mean length of selected video $V$, for the specific dataset. In this way, we can relate the  anticipation values chosen by the cited authors with our fixed anticipation value of 1 sec., which is independent of the video length. 
 Video length is shown in the first column,   anticipation time  in the second column.  Footnotes in the table  specify the number of activities in the dataset considered by the cited approach. In bold are the best in class. We can see that AFN outperforms all the approaches on anticipation, according to the displayed measure. 

\noindent
{\bf Action forecasting}
Here we compare with two methods:  \cite{yazanwill2018} on the Breakfast dataset, and \cite{mahmud2017joint} on the MPII dataset.  No works on action forecasting are based  on the  Charades and CAD120 datasets, as far as we know. 
Table \ref{tab:Bonn} confronts with  \cite{yazanwill2018}  and both Table \ref{tab:MPII1} and Table  \ref{tab:MPII2} with \cite{mahmud2017joint}.
To compare with Farha {\em et Al.} \cite{yazanwill2018} on the Breakfast dataset we  consider the $20\%$ and the $30\%$ of the video length as observation. This given we forecast over an horizon of $10\%$ of the remaining video.    The comparison takes into account both the RNN-based anticipation and the CNN-based anticipation proposed in \cite{yazanwill2018}. We have chosen these forecasting  measures, since we forecast precisely the next action, while \cite{yazanwill2018} forecast all actions up to the video end. Though this is done at the cost of an accuracy significantly lower than ours, as the table shows.

Comparisons with \cite{mahmud2017joint} are shown in Table \ref{tab:MPII1} and in Table \ref{tab:MPII2}.
In Table \ref{tab:MPII1} we compare precision and recall with respect to next action forecasting, considering only the top 1.
In Table \ref{tab:MPII2}, on the other hand, we compare the accuracy with respect to the number of past recognized actions. 
We observe that the improvement is between $11\%$ and $13\%$ percent in the first comparison and from about $6\%$ up to $17\%$ in the second comparison, as as the past sequence length increases.

\begin{figure}[t!]
 \centering
    \includegraphics[width=0.80 \linewidth]{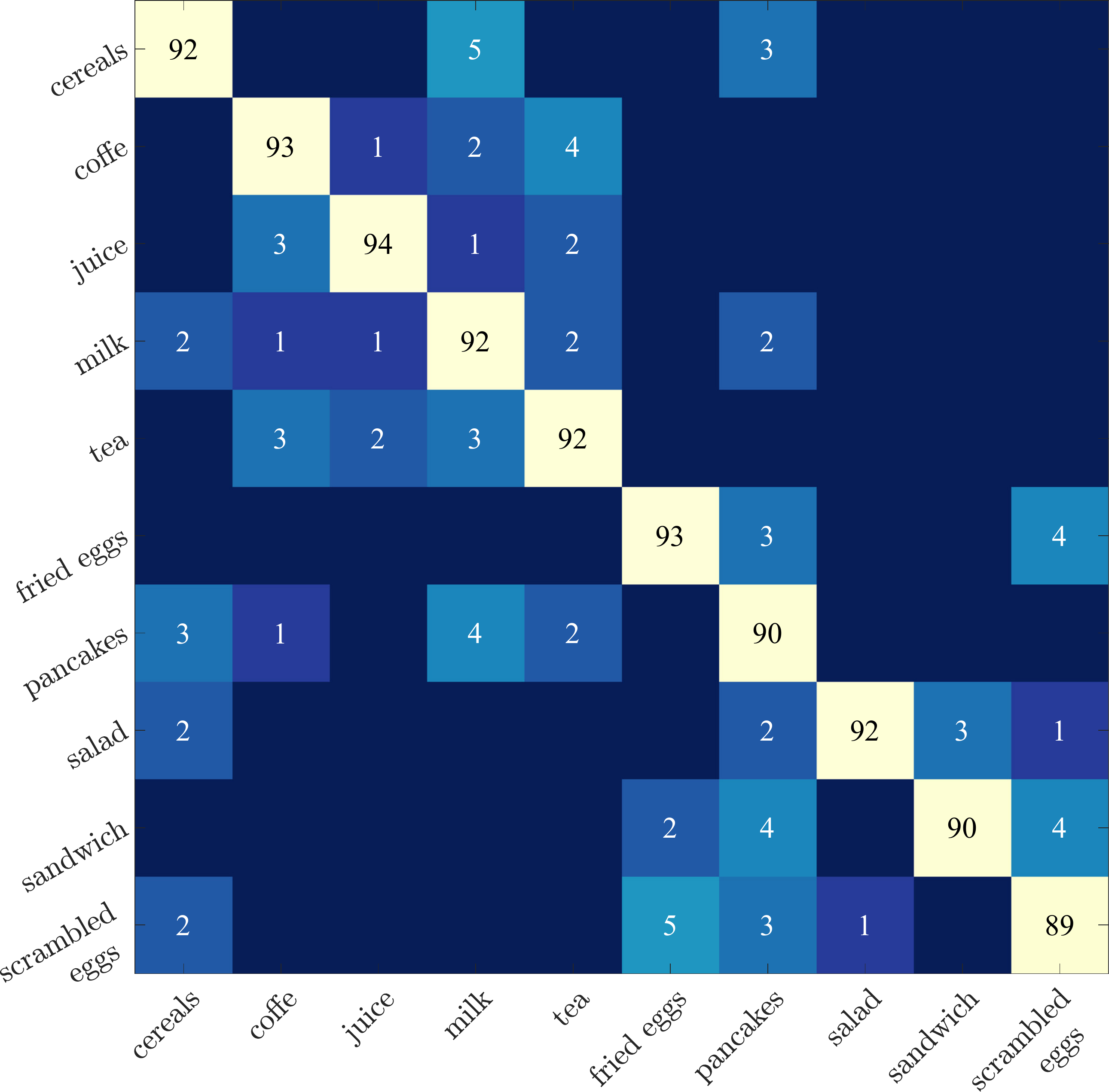}
    \caption{Confusion matrix for the Activities in Breakfast. Values less than $1$ are dropped.}\label{fig:confM}
 \end{figure}
\subsection{Evaluation of AFN}\label{ssec:eval}
We evaluate AFN specifically on next action forecasting, since for anticipation all the datasets considered are covered in Table \ref{tab:anticipation}. 
The confusion matrix shown in Figure \ref{fig:confM} shows the average accuracy of actions occurring in activities, for  Breakfast dataset. In Table \ref{tab:NAF} we show the accuracy for next action forecasting when jumping into an activity video sequence. We consider  four cases: 1) at the beginning of an action, namely when the timeline horizon to the next action is between $90\%$ and $100\%$, 2)  after the $90\%$ but before the middle of the action; 3) after the middle,  before the last $1\%$, and 4) at the end of the current action. We can see that 
the accuracy of the prediction does not significantly improve with respect to the distance from next action. On the other hand the accuracy drops to zero when the selected sample is in between the current and next action, since AFN discards these cases, see also Figure \ref{fig:tempiazioni}. The best dataset for our experiments turns out to be Breakfast.
Ablation experiments are shown in Table \ref{tab:abx}. Here, we can observe the accuracy of AFN when, in turn, the components  $q_1$, $q_4$, and the auxiliary connection to the convolutional layers of $q_0$, indicated in the table by ${\mathcal L}_3$, are removed.
  
\begin{table}[h!]
\caption{Evaluation of AFN for next action forecasting}\label{tab:NAF}
\resizebox{\columnwidth}{!}{
\begin{tabular}{|l|l|l|l|l|}
\hline
Dataset &\multicolumn{4}{|c|}{\textbf{Accuracy $\%$  according to timeline horizon $h\%$ to next action} } \\ \hline
& \textbf{90\%$\leq$h$<$100\%}  & \textbf{50\%$\leq$h$<$90\%} & \textbf{1\%$\leq$ h$<$50\%} &  \textbf{0\%$\leq$ h$<$1\%} 
\\ \hline
MPII \cite{rohrbach2012database} & 93.4\%             & 93.7\%    & 93.9\%        & 0\%     
\\ \hline
 Breakfast \cite{Kuehne12} & \textbf{94.2\%}    & \textbf{94.6\%} & \textbf{94.72\%} &0\%
\\ \hline   
Charades \cite{sigurdsson2016hollywood}  & 89.5\%    & 89.8\% & 90.11\% & 0\%
\\ \hline
CAD120 \cite{koppula2013dataset} & 92.6\%    & 93.2\% & 93.5\% & 0\%
\\ \hline                                                 
\end{tabular}
}
\end{table}

More results are shown in the supplementary materials.

\end{document}


\title{Supplementary Material}

\author{}
\date{}
\maketitle

\graphicspath{{figssupp/}}

\section{Next action label}\label{sec:intro}
In this supplementary material we provide an evaluation for the proposed next action forecasting framework. In particular, we present confusion matrices for the datasets Charades, Breakfast, CAD120 and MPII-kitchen. We also provide an  accuracy according to the time distance between the anticipation of the current action and the starting time of the next action (see also Table 6 in the paper). To produce the confusion matrices we specifically recorded the false positives. For true and false negatives we considered the ``action'' {\em null} in CAD120, the  {\em background} in the  dataset MPII-Kitchen,  denoting  when no activity is detected. Similarly  for the dataset Breakfast we considered the {\em SIL} action specifying the start and end of an action. 

The {\em time to next action} accuracy graph  shows  the time left to the next action start. To illustrate this accuracy value we have selected actions from the datasets with highest time variance defined as follows. Let $S$ be the set of reference (the test set), $A(t)$ be the label of an  action,
 and let $\Delta A(t)= A'(t)-A(t^+)$, with $A'$ a label of any of the actions following $A$ in the set considered, and $(t^+)$ the second at which $A$ is anticipated. Let $\mu_A = \frac{1}{N}\sum_{i=1}^N \Delta A(t_i)$ with $A(t_i)$ the same action $A$ with $t_i$ varying in all occurrence of $A\in S$. We say that $A$ has a high variation (with respect to forecasting and anticipation) if $\mu_{A}/\mu_{B}>\lambda$ for some $\lambda$ experimentally established for each set, and for a number of other actions $B$ in $S$ (where the number depends on $\lambda$). When $\Delta A(t)$ is negative then there is no next action $A'$. This variation  is relevant because it shows if  accuracy in forecasting is  independent of the subject performing an action and the circumstances in which it is performed.
All evaluations are done on the test sets.

\subsection{CAD120 Dataset}
CAD120 dataset \cite{koppula2013dataset} has 120 videos annotated with 10 actions and 10 high-level activities. 
 Note that the actions are the same for all the activities.  This is particularly interesting for the ProtoNet component of AFN, thus we show  also the confusion matrix for the activities of CAD120.

\begin{figure*}
 \centering
    \includegraphics[width=0.43 \linewidth]{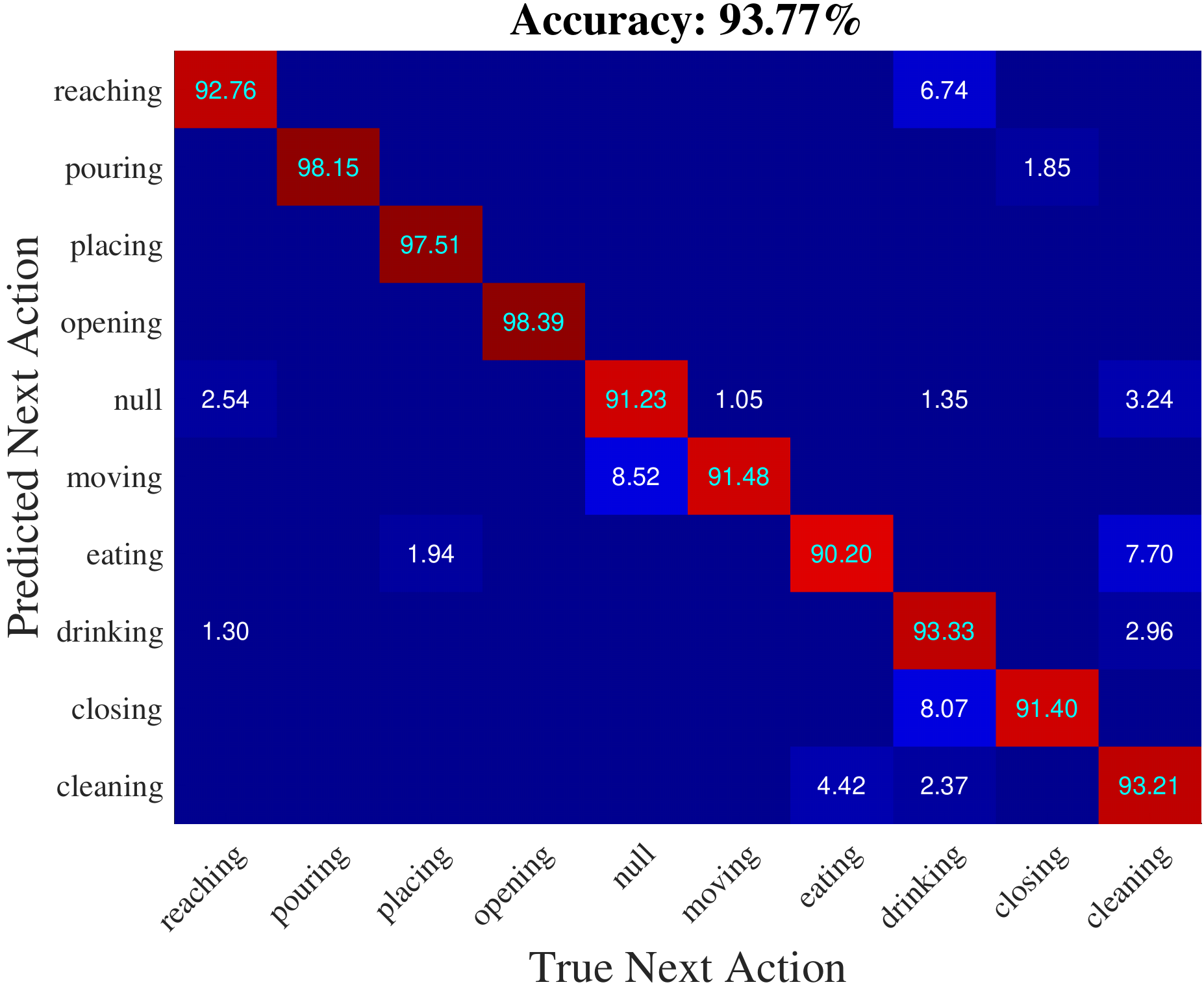}
    \includegraphics[width=0.53 \linewidth]{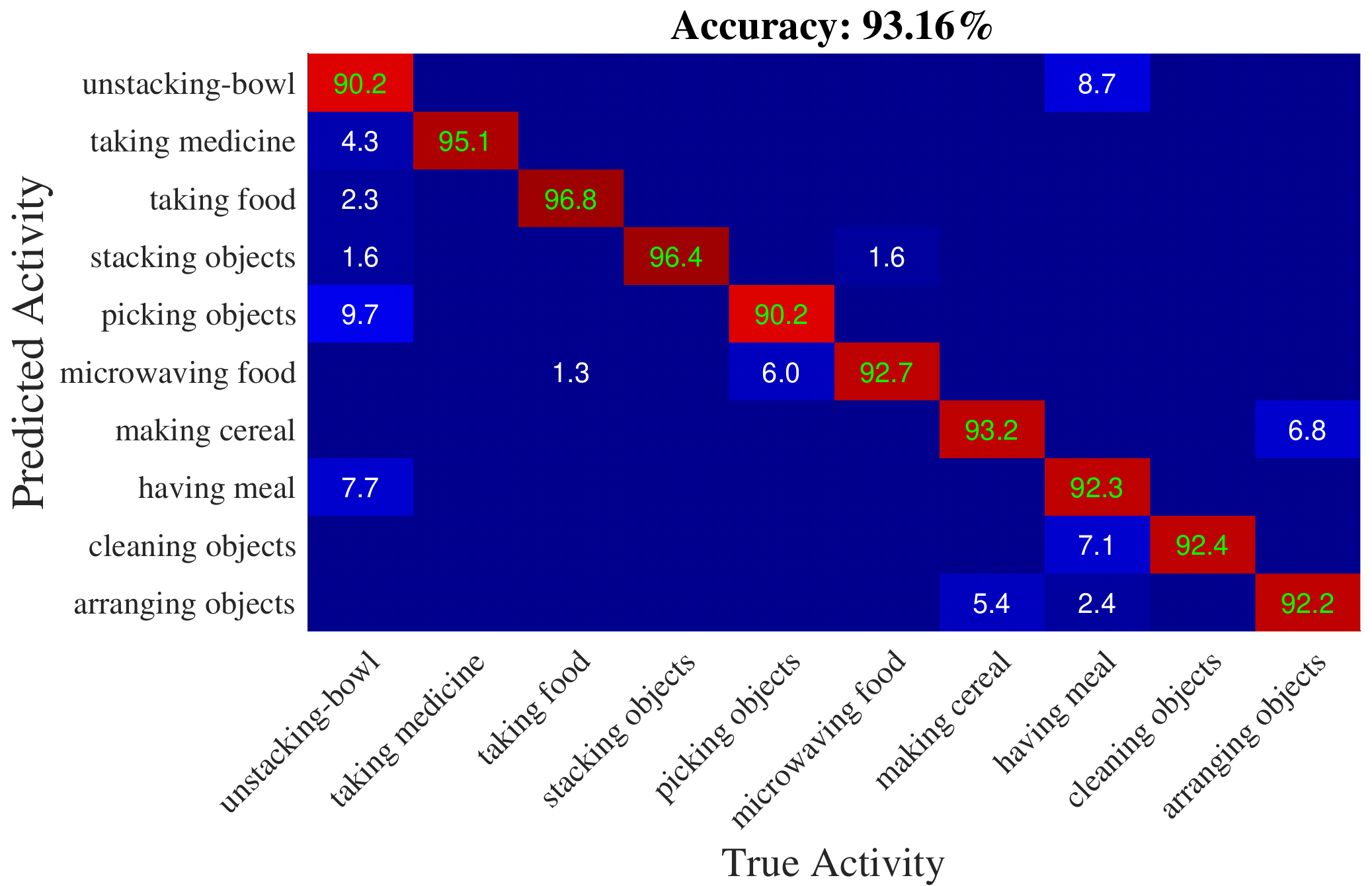}
    \caption{On the left the Confusion Matrix for  CAD120 dataset. Values less than 0.8 are not shown, the heat  color-map is {\em jet}. The {\em null} action is recorded to show false negatives. On the right Confusion Matrix for CAD120 Activities.  }\label{fig:confCADActivities}\label{fig:confCAD}
 \end{figure*}

\begin{figure*}
 \centering
    \includegraphics[width=0.98 \linewidth]{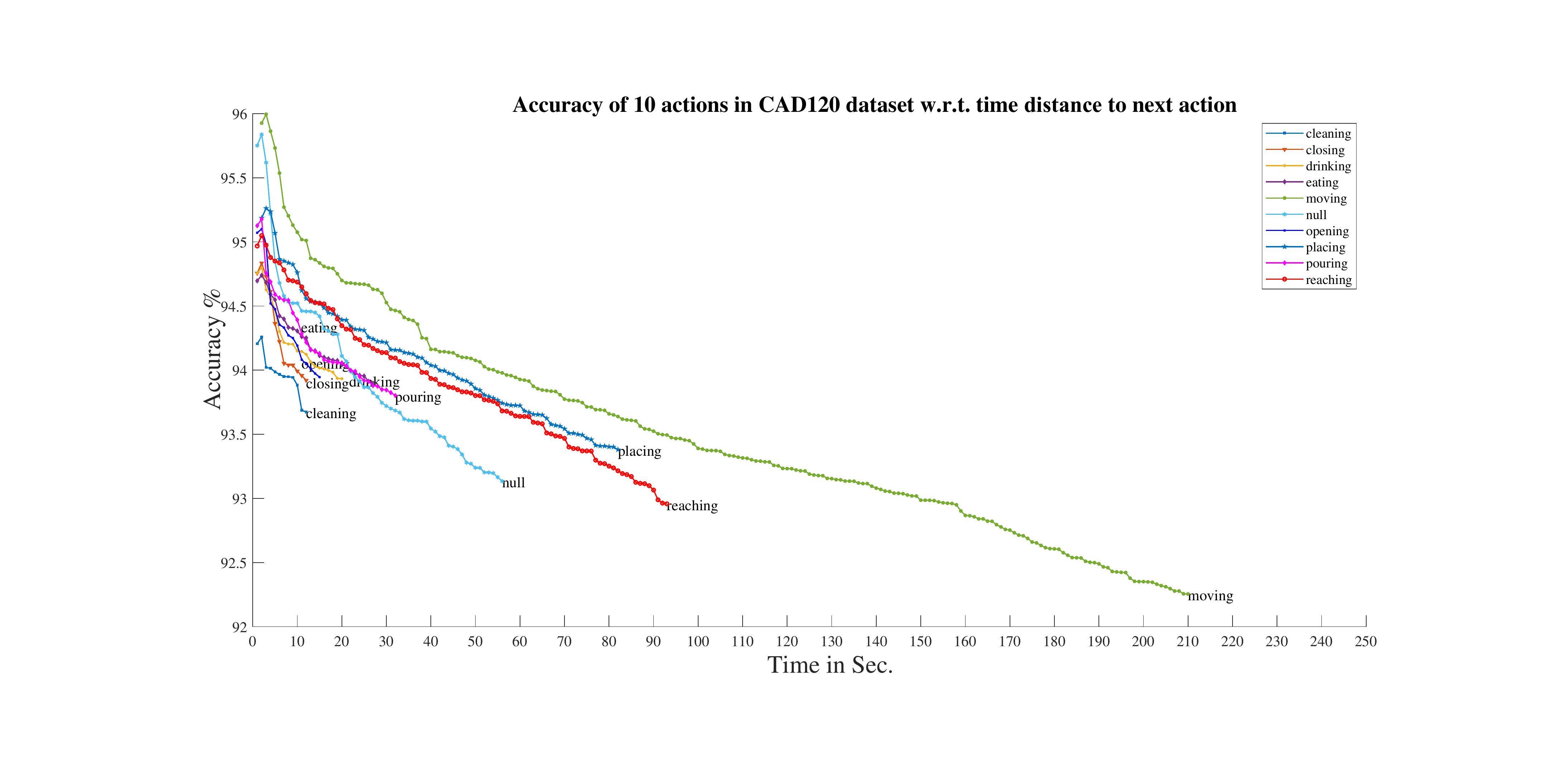}
    \caption{Accuracy with respect to time distance between  current action anticipation  to next action start for dataset CAD120. Time is computed according to videos framerate.} \label{fig:CAD-time}
 \end{figure*}

Figure \ref{fig:confCAD} shows on the left the confusion matrix, including the {\em null} action and on the right it shows the classification by the ProtoNet component of AFN.  Figure \ref{fig:CAD-time} shows the next action forecasting accuracy with respect to the time distance between the current action anticipation and the next action start. Here we considered all the actions.

\subsection{MPII-cooking dataset}
The MPII cooking activities \cite{rohrbach2012database} dataset consists of 65 actions recorded from 12 subjects. 

\begin{figure*}
 \centering
    \includegraphics[width=0.98 \linewidth]{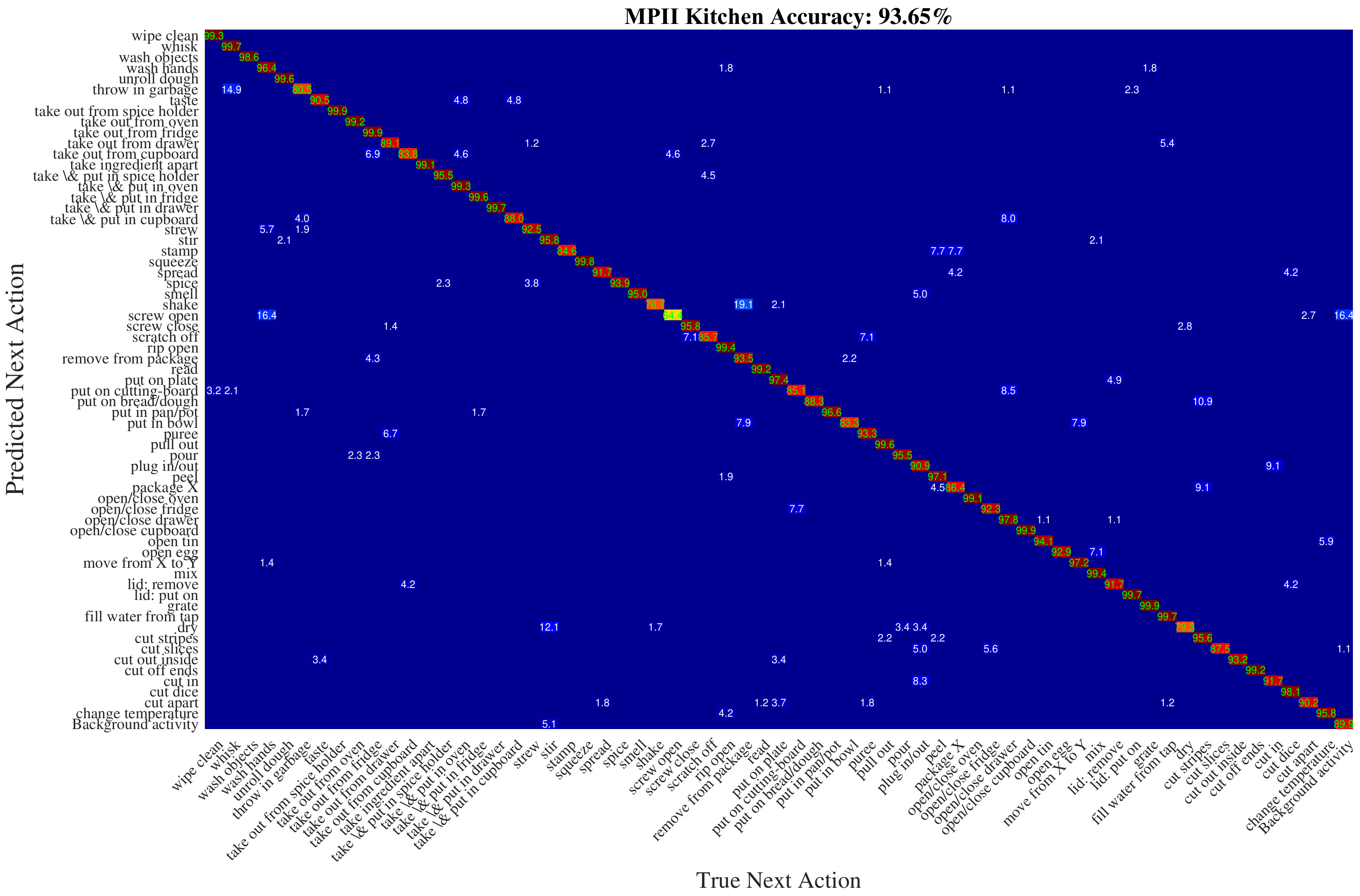}
    \caption{Confusion matrix for the MPII-kitchen dataset. The heat-map is {\em jet}, values less than $1$ are not shown.} \label{fig:confMPII}
 \end{figure*}

For this dataset there are 44 videos at high resolution, with a framerate of 29.4hz. The confusion matrix for the next activity prediction is illustrated in Figure \ref{fig:confMPII}. True positive are illustrated with a jet map and small digits indicate the accuracy of correct predictions. In white the false positive. To record false negatives we used the fictitious action {\em background} included in the dataset annotations.

\begin{figure*}
 \centering
    \includegraphics[width=0.98 \linewidth]{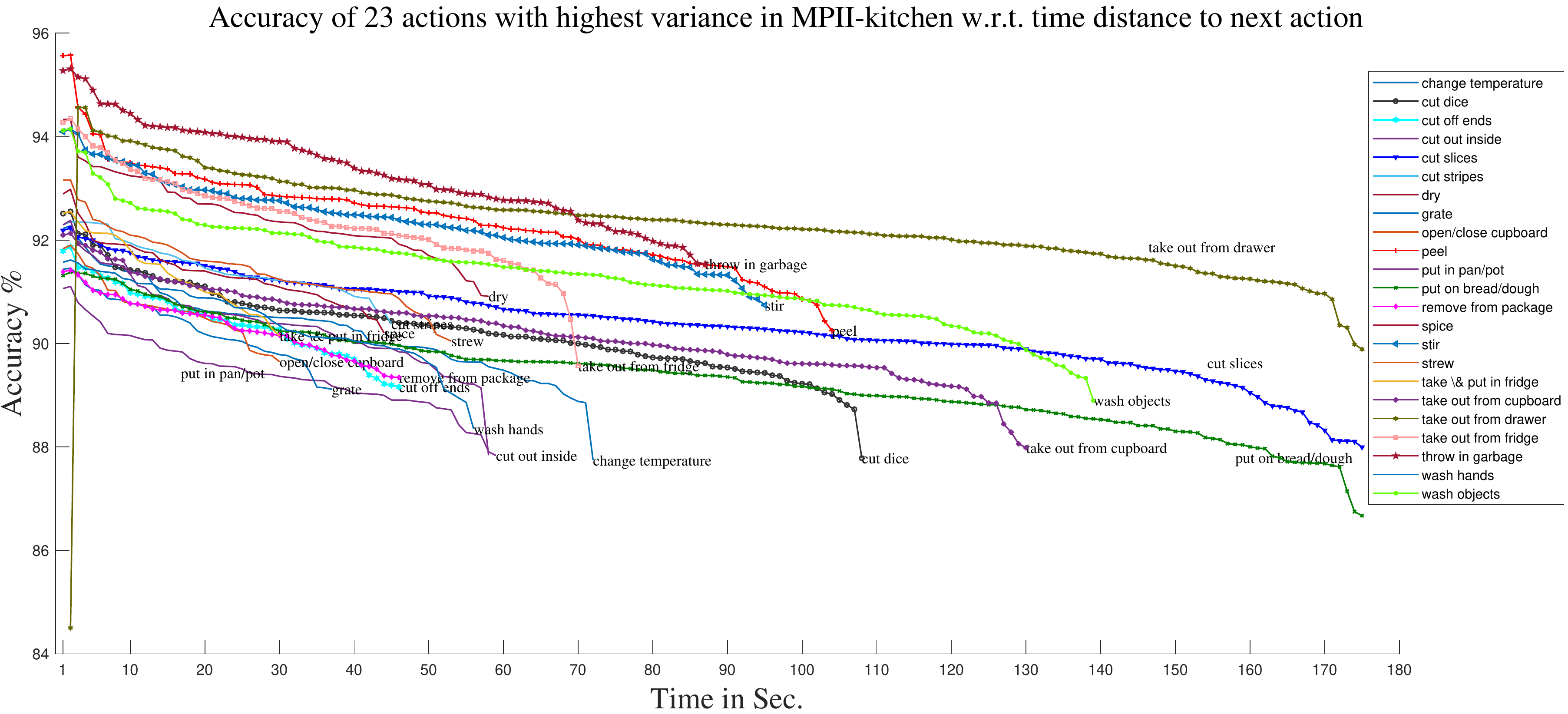}
    \caption{Accuracy trends for next action prediction for MPII-kitchen dataset, considering the 23  actions with greatest variance over the 64 in the dataset. The time line indicates the time distance to the next action, for each occurrence of the indicated action, in the video dataset. Framerate is 29.4Hz} \label{fig:time-MPII}
 \end{figure*}

Figure \ref{fig:time-MPII} shows the accuracy of  next action forecasting according to the time distance between the first second of the current action to the start point of the next action. We considered the 23 actions, of the 65 ones, with highest variation for $\lambda=2$.
Videos in MPII dataset are acquired at 29.4Hz. According to the framerate   there is a  $1\%$ of cases in which the time lapse between the current-action-start and the  next-action-start is less than a second, this $1\%$ is collected as an error for AFN.

\subsection{Breakfast Dataset}
\begin{figure*}
 \centering
    \includegraphics[width=0.98 \linewidth]{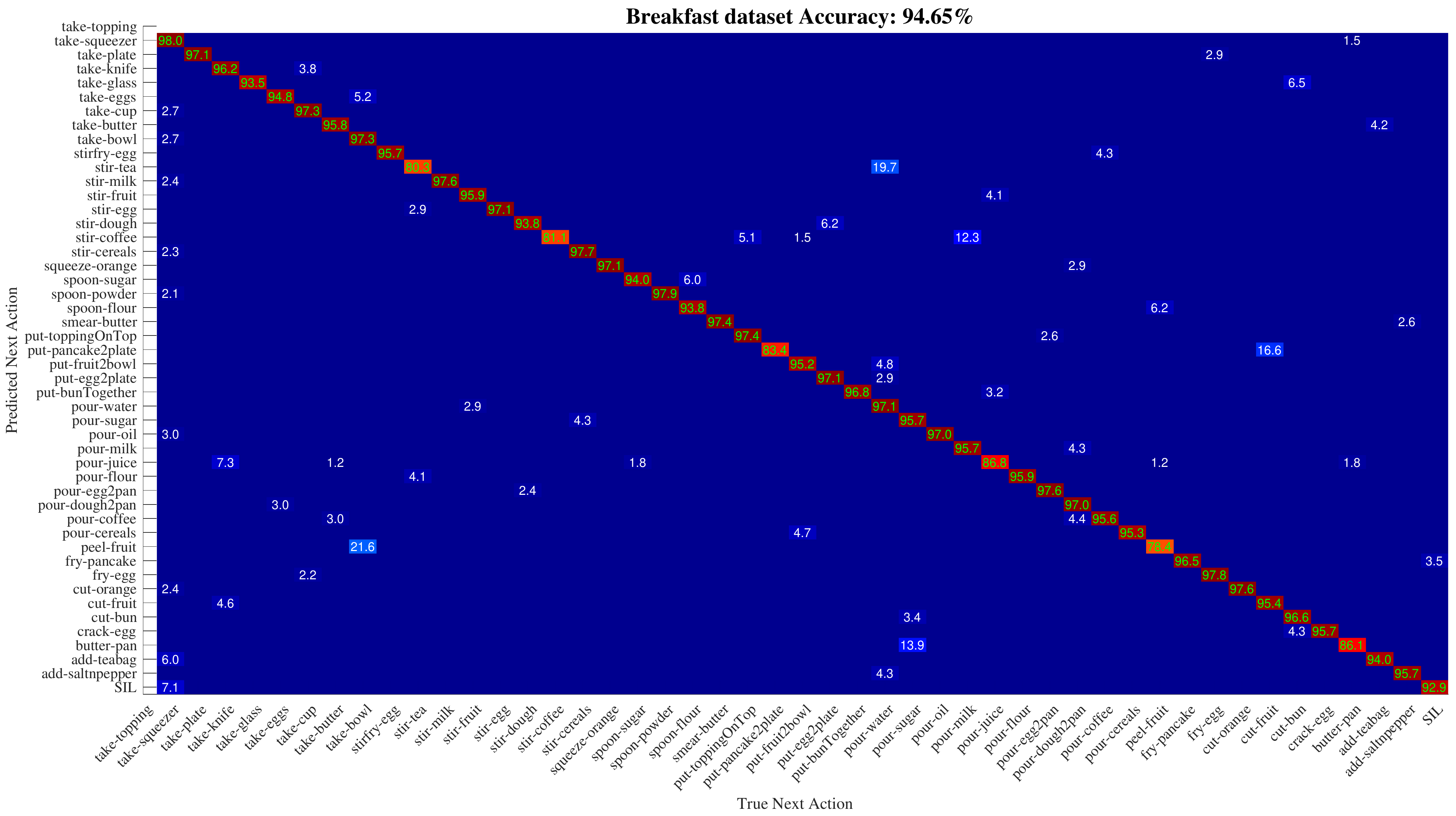}
    \caption{Confusion matrix for the Breakfast dataset. The heat-map is {\em jet}, values less than $1$ are not shown. The {\em SIL} actions is recorded for false negatives.} \label{fig:ConMatBreakfast}
 \end{figure*}

The Breakfast dataset \cite{Kuehne12}, in its rough version has 48 actions occurring in 10 activities (see Figure 6 in the submitted paper). Figure \ref{fig:ConMatBreakfast} shows the complete confusion matrix, including the {\em SIL} operator, considered for recording false negative. For breakfast it amounts to forecast the existence of a next action, while the activity should have been concluded.

\begin{figure*}
 \centering
    \includegraphics[width=0.98 \linewidth]{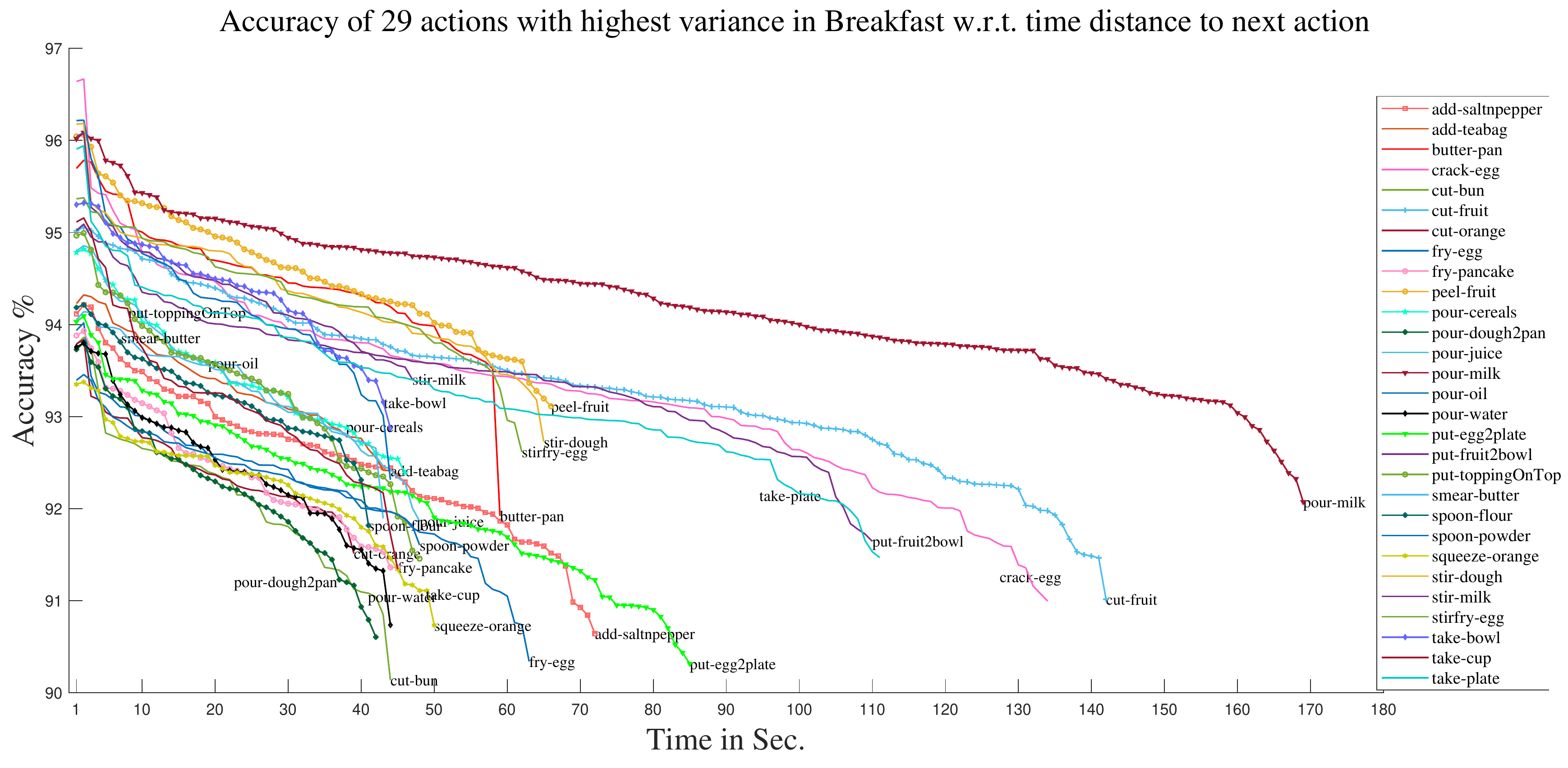}
    \caption{Accuracy trends for next action prediction, for the Breakfast dataset. The time line indicates the time distance to the next action, for each occurrence of the indicated action, in the video dataset.} \label{fig:time-Breakfast}
 \end{figure*}

In Breakfast dataset, framerate is 15Hz, and the time lapse between the current action anticipation and the next action start has a mean value of $5.1$ seconds. On the other hand  a less than a second time lapse between the current action start and the next action start occurs only in $4\%$ of all the videos (77 hours of videos). We consider this amount an error for our framework. As we did for MPII-kitchen also in this case we considered the actions with highest variance to evaluate the accuracy on the time distance with respect to the next action start, here we choose $\lambda=0.82$. Accuracy for time distance is illustrated in Figure \ref{fig:time-Breakfast}.

\subsection{Charades Dataset}
\begin{figure*}
 \centering
    \includegraphics[width=0.98 \linewidth]{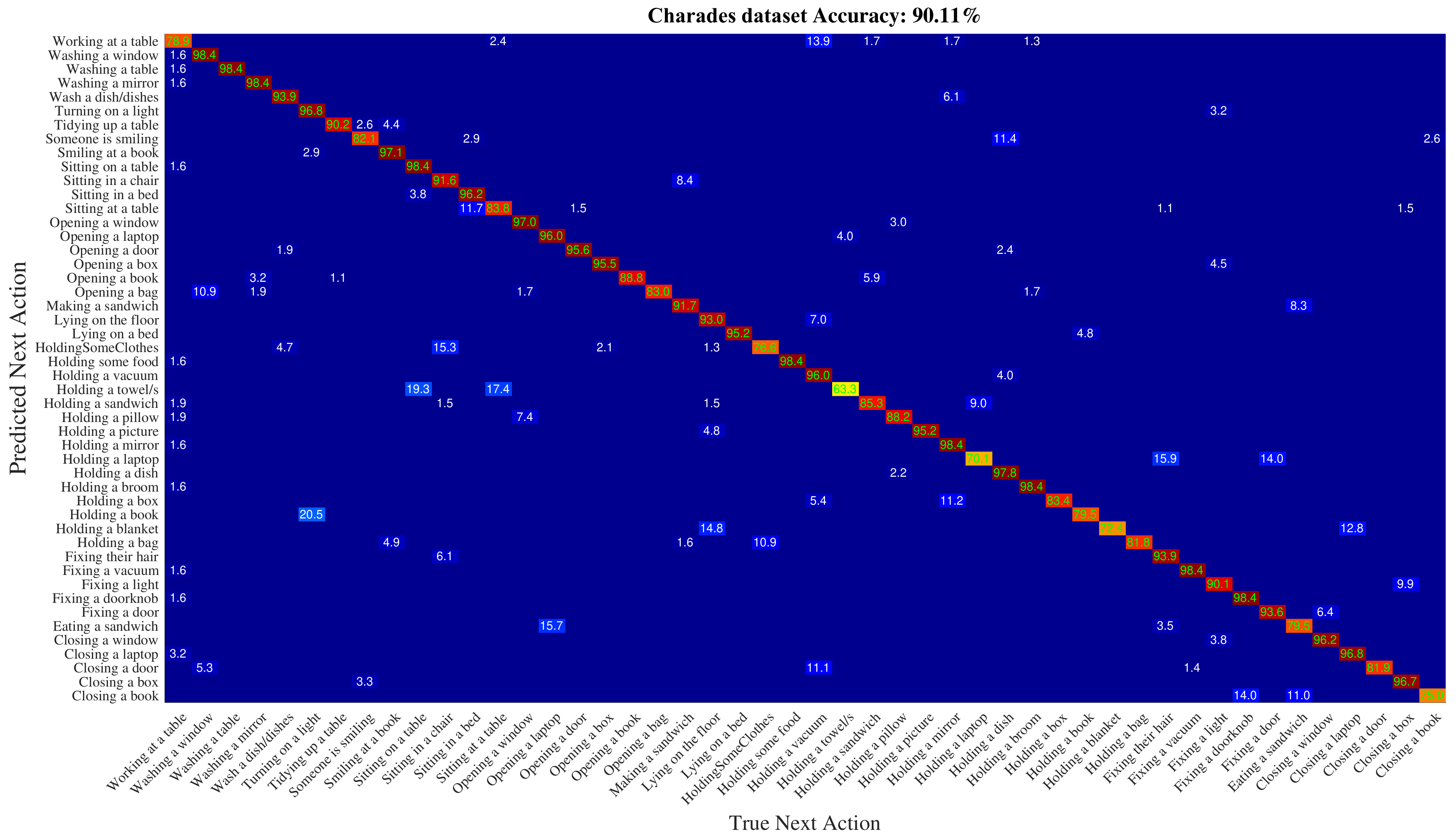}
    \caption{Confusion matrix for the CHARADES dataset. The heat-map is {\em jet}, values less than $1$ are not shown. } \label{fig:ConfMatCharades}
 \end{figure*}

The Charades dataset \cite{sigurdsson2018} is a very rich dataset with 157 action classes, objects, affordances and descriptions. Because all actions are specified with respect to a context scene we considered the context scenes as the high level activities for the ProtoNet. Each video has on average 6.8 actions and several actions are considered co-occurring, such as closing-opening a window, so to obtain a sequence we had to adapt  the time intervals. For Charades we used the proposed separation in training and test. Figure \ref{fig:ConfMatCharades} shows the confusion matrix for a subset of 48 actions for visibility reasons (chosen randomly over the set of 156 actions). For an action sequence completion we have added an {\em end}, not shown in the confusion matrix. 

\begin{figure*}
 \centering
    \includegraphics[width=0.98 \linewidth]{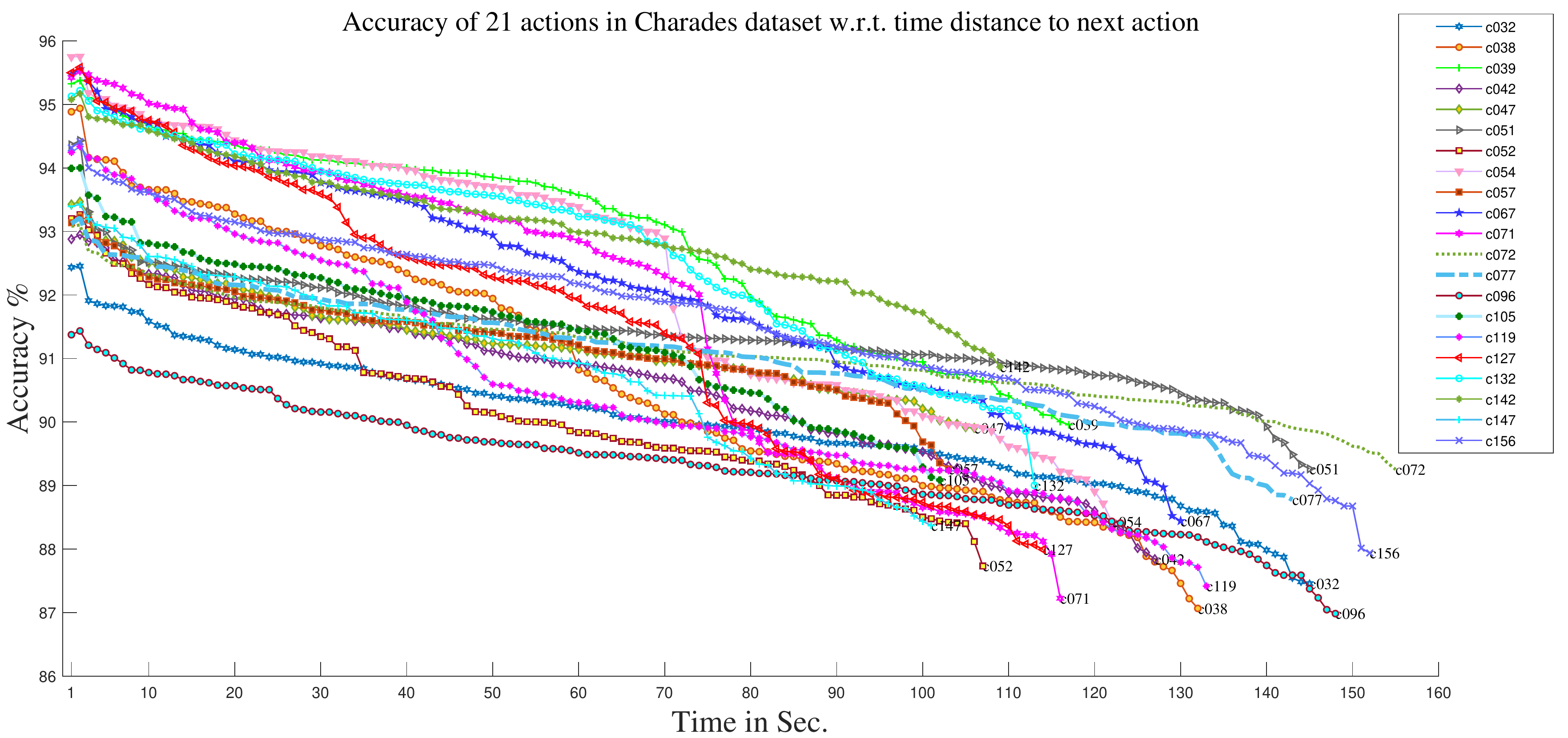}
    \caption{Accuracy trends for next action forecasting, in CHARADES dataset. The codes for actions are those used in CHARADES since action names are often very long. Here the codes are used to avoid clumsiness.} \label{fig:time-Charades}
 \end{figure*}

Similarly, we have chosen a $\lambda = 0.94$  to analyze the accuracy with respect to time distance to next action, illustrated in Figure \ref{fig:time-Charades}.

\typeout{*********************** MPII Dataset**********************}

\typeout{*********************** MPII Dataset**********************}

 \typeout{*************** Breakfast dataset***************}

\typeout{*************** Charades***************}

\typeout{*************** biblio***************}